\pdfoutput=1

\documentclass[11pt]{article}

\usepackage[final]{acl}

\usepackage{times}
\usepackage{latexsym}

\usepackage[T1]{fontenc}

\usepackage[utf8]{inputenc}

\usepackage{microtype}

\usepackage{inconsolata}

\usepackage{graphicx}

\newcommand{\hidden}[1]{}
\newcommand{\arwi}{\textsc{Arwi}}
\title{{\arwi}: Arabic Write and Improve}

\author{Kirill Chirkunov,\textsuperscript{1} Bashar Alhafni,\textsuperscript{1,2} Chatrine Qwaider,\textsuperscript{1} \\ \textbf{Nizar Habash,\textsuperscript{1,2}} \textbf{Ted Briscoe\textsuperscript{1}}\\
  \textsuperscript{1}MBZUAI, \textsuperscript{2}New York University Abu Dhabi\\
  \texttt{\{kirill.chirkunov,chatrine.qwaider,ted.briscoe\}@mbzuai.ac.ae}\\
  \texttt{\{alhafni,nizar.habash\}@nyu.edu}
  }

\begin{document}
\maketitle
\begin{abstract}

Although Arabic is spoken by over 400 million people, advanced Arabic writing assistance tools remain limited. To address this gap, we present {\arwi}, a new writing assistant that helps learners improve essay writing in Modern Standard Arabic. {\arwi} is the first publicly available\footnote{\url{https://arwi.mbzuai.ac.ae/}} Arabic writing assistant to include a prompt database for different proficiency levels, an Arabic text editor, state-of-the-art grammatical error detection and correction, and automated essay scoring aligned with the Common European Framework of Reference standards for language attainment. Moreover, {\arwi} can be used to gather a growing auto-annotated corpus, facilitating further research on Arabic grammar correction and essay scoring, as well as profiling patterns of errors made by native speakers and non-native learners.
A preliminary user study shows that {\arwi} provides actionable feedback, helping learners identify grammatical gaps, assess language proficiency, and guide improvement.

\end{abstract}

\section{Introduction}


Arabic is the national language of over 400 million people and one of the UN’s six official languages \cite{Ryding_Wilmsen_2021,UNPage2024}. Yet, Arabic writing assistance tools remain severely underdeveloped. Unlike English, which has numerous competitive writing assistants and CEFR-benchmarked grading systems \cite{cefr2001}, Arabic tools are limited to a few commercial error-correction systems with no objective public evaluation. Enhanced writing assistants could benefit millions of Arabic writers and aid corpus collection, advancing Arabic NLP.



The development of Arabic writing assistants faces major challenges, with one of the most significant being the lack of a diverse Arabic corpus that captures the wide range of writing variations, including grammatical errors made by both native speakers and second language learners. Having such a comprehensive corpus would enable the creation of writing assistants that not only provide accurate error detection and correction suggestions but also motivate learners to continuously enhance their Arabic writing skills. Additionally, these assistants would contribute to ongoing data collection while actively supporting users in refining their writing abilities.

In response to these challenges, we introduce \textbf{{\arwi}}, a writing assistant tool specifically designed to help MSA writers improve their essay-writing skills. \textbf{{\arwi}} features an intuitive interface and user experience based on the following core components:

\begin{itemize}
 







    \item \textbf{Essay Prompt Database}: A library of writing topics across CEFR levels.  
    \item \textbf{Arabic Text Editor}: Highlights errors, aids structuring, and supports iterative drafting.  
    \item \textbf{Grammar Error Detection \& Correction (GED/C)}: Identifies errors (e.g., orthography, morphology) and offers feedback.  
    \item \textbf{Automated Essay Scoring (AES)}: Assesses grammar, vocabulary, and errors to estimate CEFR levels (A1-C2).  
    \item \textbf{Progress Tracking}: Stores revisions and visualizes improvement.  
    \item \textbf{User Profiling}: Allows learners to specify dialect, native language, and proficiency.  
    \item \textbf{Auto-Annotated Corpora}: A growing repository of diverse, auto-annotated essay samples.  

\end{itemize}

Section \ref{sec:related-word} presents related work; and Section \ref{sec:system} presents a description of the {\arwi} system. We discuss a preliminary user experiment in Section \ref{sec:user-study}, and our conclusions and outlook in Section \ref{sec:conclusion}.

\section{Related Work}
\label{sec:related-word}
\subsection{Existing datasets for writing improvement}

Prominent English datasets include the CoNLL-2014 corpus \cite{ng-etal-2014-conll}--derived from the NUCLE \cite{dahlmeier-etal-2013-building} release with approximately 1.2 million words--along with WI-LOCNESS \cite{bryant-etal-2019-bea,granger1998} which offers 3,000 annotated essays (628K words) grouped by CEFR levels. More recently, the Write \& Improve annotated corpus \cite{wicorpus24} has provided a large resource of 23,000 annotated essays with detailed CEFR annotations, supporting both Grammatical Error Detection/Correction (GED/C) and Automatic Essay Scoring (AES) tasks. In addition, several English GED/C datasets such as GMEG-Yahoo and GMEG-Wiki \cite{napoles-etal-2019-enabling} extend the scope by covering different business domains as well as formal and informal speech registers. The JFLEG dataset \cite{napoles-sakaguchi-tetreault:2017:EACLshort} further complements these resources by focusing on fluency as opposed to minimal meaning-preserving edits.

Arabic datasets are limited in both size and diversity. The QALB-2014 corpus \cite{mohit-etal-2014-first} contains around 1.2 million words across 21,396 sentences from online commentaries on Al Jazeera articles, each paired with a corrected version to facilitate GED/C research. QALB-2015 \cite{rozovskaya-etal-2015-second} adds another layer by offering 622 annotated essay sentences (approximately 140K words) from both native and non-native writers. Complementing these, the ZAEBUC corpus \cite{habash-palfreyman-2022-zaebuc} comprises 214 annotated Arabic essays (about 33.3K words) with CEFR grades, thus addressing both GED/C and AES tasks. However, even combined, these Arabic resources lack the extensive genre, topic and proficiency-level stratification of their English counterparts. 

\subsection{Arabic Writing Assistance Tools}


In contrast to numerous English writing assistants like Write\&Improve,\footnote{\url{https://writeandimprove.com/}} Grammarly, and others \cite{writing2024tools}, which assess fluency and grammar, Arabic tools (e.g., Sahehly,\footnote{\url{https://sahehly.com/}} Qalam\footnote{\url{https://qalam.ai/}}) focus on common errors but lack overall writing quality feedback.  
They show good performance in identifying and correcting common errors, such as Hamza placement or confusion between Ha, Ta, and Ta-Marbuta, but lack the capability to detect and correct more nuanced error types, such as merge/split errors or issues related to the shortening of long vowels, as outlined in comprehensive error taxonomies \cite{alfaifi2012arabic,alfaifi2013error}.

\subsection{LLMs as Arabic Writing Assistants}


The advent of large language models (LLMs) has led to the development of writing assistants based on zero-shot or few-shot prompt engineering \cite{fitria2023artificial,Yancey2023RatingSL,pack2024large,kim2024designingpromptanalyticsdashboards}, as seen in multilingual (ChatGPT, Gemini, etc.) and Arabic-centric LLMs (Jais Chat \cite{sengupta2023jais} and Fanar \cite{team2025fanar}). Despite their strong baseline performance, these models tend to fall short when compared to specialized systems focused on GED/C and AES \cite{wu2023chatgptgrammarlyevaluatingchatgpt,alhafni:2025:enhancing}.

Recent fine-tuning experiments on English GED/C and AES datasets have yielded promising results, demonstrating that pretrained LLMs can achieve state-of-the-art performance in GEC \cite{omelianchuk2024pillarsgrammaticalerrorcorrection} if used within ensemble models. This observation underscores the potential benefits of creating a rich, diverse corpus of annotated Arabic texts, which would facilitate the fine-tuning of LLMs specifically for MSA writing assistance.

\begin{figure*}
    \centering
    \includegraphics[width=1\linewidth]{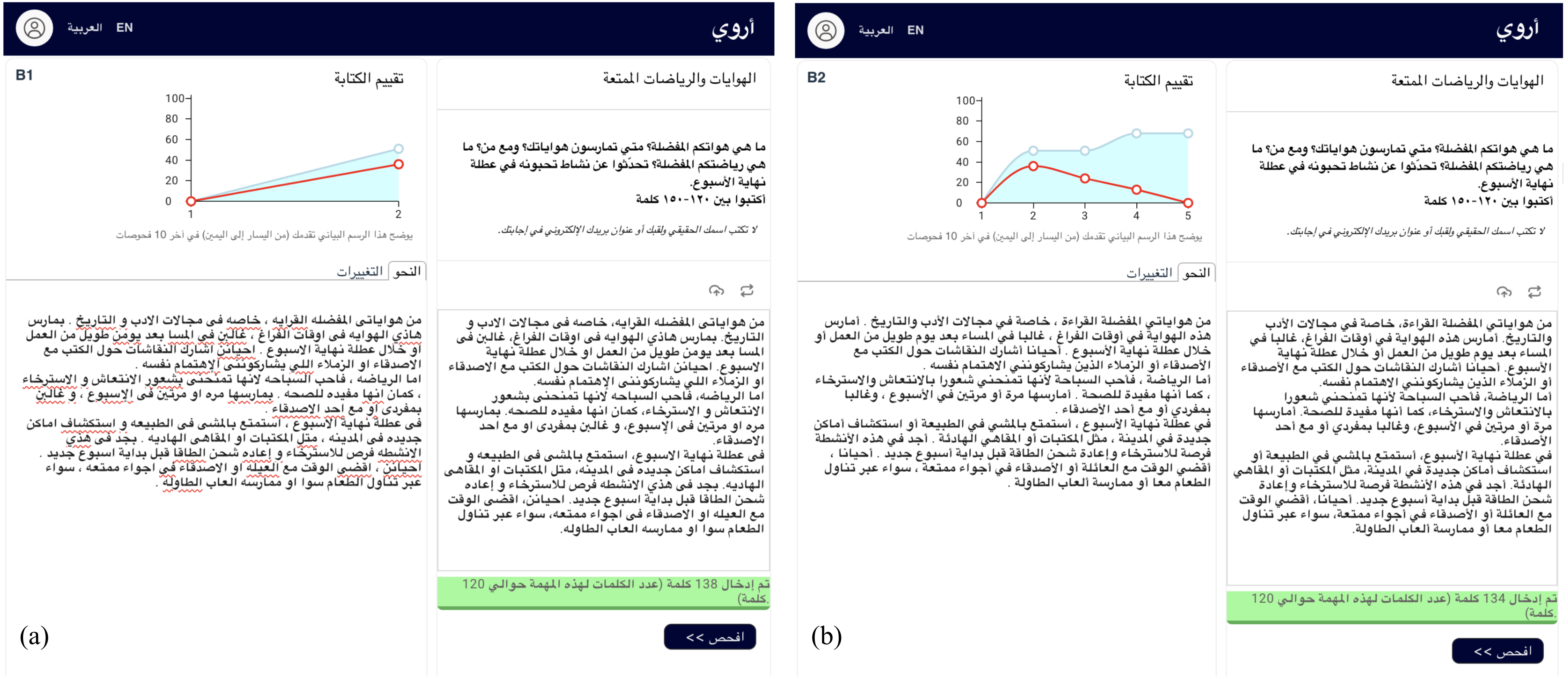}
    \caption{A before-and-after example of using  {\arwi}'s Arabic interface. In (a) the text receives a B1 CEFR and a large number of errors marked with red underlining; in (b) the results shows improved writing and is raised to B2 CEFR. See Appendix~\ref{appendix} for the English version of the interface. }
    \label{fig:arwi-interface}
\end{figure*}

\section{System Description}
\label{sec:system}
\subsection{Overview of {\arwi}}
{\arwi} functions as a web application, integrating a front-end user interface with a backend of specialized REST API services and data collection infrastructure. The system includes an Arabic text editor with diacritics support, GED/C auto-annotation, AES module, and progress tracking that displays learning trajectories and revision improvements.  {\arwi} delivers personalized, actionable feedback to help users continuously enhance their writing skills. 
Screenshots of the system are provided in Figure~\ref{fig:arwi-interface} 
to illustrate {\arwi}'s current UI/UX and typical pattern of use. Figure \ref{fig:arwi-interface-en} in Appendix \ref{appendix} shows the English version of the interface.

\subsection{Core Components}

\subsubsection{Collection of Essay Prompts}


We develop an expandable database of essay prompts to provide targeted writing tasks for all CEFR levels. Each prompt covers a specific topic across various domains, aligning with Arabic cultural sensitivities and supporting both formal and informal genres. {\arwi} enforces a minimum word count: 50 words for beginners (A1-A2), 100 for intermediate writers (B1-B2), and 200+ for advanced learners (C1-C2).


Beginner prompts focus on descriptive writing (e.g., favorite animals, family traditions). Intermediate learners engage with reflective or argumentative topics (e.g., pros and cons of wearing uniforms), while advanced writers tackle analytical discussions (e.g., AI ethics, environmental sustainability). Additionally, some prompts include optional media elements, such as images, to support descriptive tasks involving processes, interior spaces, or graphical representations.



Many Arabic proficiency exams, including CIMA\footnote{\url{https://www.imarabe.org/}} and ALPT\footnote{\url{https://www.arabacademy.com/alpt/}}, require writing tasks. Our essay prompt design draws inspiration from these exams, aligning with their task types. By mapping prompts to the CEFR scale, we ensure appropriate difficulty levels and help learners prepare for CEFR-benchmarked Arabic proficiency tests.

\hidden{
\paragraph{Arabic CEFR}
Several efforts have applied CEFR leveling to Arabic texts at various granularities.
The KELLY project \cite{Kilgarriff:2014:corpus-based} mapped 9,000 frequent words across nine languages, including Arabic, to CEFR levels using corpus analysis.
\newcite{Habash:2022:zaebuc} manually annotated Arabic and English essays.
\newcite{AboAmsha2022:ArabicCEFR} provided a reference on Arabic CEFR leveling for non-native speakers.
\newcite{naous_readme_2023} created a CEFR-leveled dataset in five languages, including Arabic.
\newcite{soliman2024creating} developed an Arabic vocabulary profile for CEFR A1–A2 based on dialectal prevalence, frequency, and complexity.
}

\subsubsection{Arabic Text Editor}


The editor disables \textit{real-time} spell-checking and auto-corrections, instead providing actionable feedback from the GED/C module upon submission. This approach encourages users to review and apply changes manually, reinforcing learning and improving retention. 
See Figure~\ref{fig:arwi-interface}.

\hidden{
Additionally, the editor includes clear warnings advising users against including sensitive or personal information in their submissions, as the annotated text may be incorporated into publicly available GEC/AES corpora. By combining a robust input interface with integrated navigational support from the GED/GEC feedback system, the editor not only facilitates high-quality text production but also fosters a proactive learning process that motivates users to refine their writing skills through deliberate engagement.
}

\hidden{
\begin{itemize}
  \item \textbf{Mechanism:} Describe how the system identifies errors (e.g., rules, machine learning, morphological analysis).
  \item \textbf{Feedback Presentation:} Explain how errors are classified (orthography, morphology) and how suggestions are displayed.
  \item \textbf{Examples:} Provide a brief snippet illustrating error detection and correction steps.
\end{itemize}
}

\subsubsection{GED/C Module}
For GED, we adopt a two-stage token-level classification approach, similar to \citep{alhafni-etal-2023-advancements}, by fine-tuning CAMeLBERT-MSA \cite{inoue-etal-2021-interplay}. The first classifier performs binary GED, identifying whether a token is erroneous, while the second classifier provides a more fine-grained analysis, categorizing errors based on the ARETA taxonomy \cite{belkebir-habash-2021-automatic}. These classifiers are applied sequentially: the binary classifier runs first, followed by the fine-grained classifier. This cascaded setup ensures high precision in our GED module.

For GEC, we develop a text-editing system that predicts character-level edits for each input token, generating the corrected text when applied \cite{alhafni:2025:enhancing}. Both GED/C models are fine-tuned on a combination of QALB-2014 and ZAEBUC.

\subsubsection{AES Module}

The AES module leverages a fine-tuned version of CAMeLBERT-MSA  to predict the CEFR levels of MSA essays. We fine-tune CAMeLBERT-MSA was on the ZAEBUC dataset and a larger synthetic dataset with topic, genre and level diversity for essay scoring \cite{qwaider2025enhancingarabicautomatedessay}.

\subsubsection{User Progress Tracking System}
The User Progress Tracking System provides writers with clear and measurable feedback, recording CEFR scores and tracking error reduction over time. This historical data is presented through a linear graph that dynamically illustrates the user's learning trajectory.

\subsubsection{User Profiling}

Users who register have the option to input their native language or Arabic dialect and estimated proficiency level. This metadata enables more targeted prompting and feedback. It also supports further annotation of the auto-annotated essays collected to create, for example, (non-)native, dialect, or CEFR level specific profiles of users.

\section{Preliminary User Experiment}
\label{sec:user-study}
Our goal is to determine if {\arwi}'s feedback leads to measurable improvements in text quality such as reduction in grammatical errors or increased CEFR scores, and whether users find the UI/UX intuitive.

\subsection{Experimental Setup}

A total of 34 non-native mixed-gender undergraduate Arabic learners organized into four groups participated with proficiency levels ranging from A1-B1. 
Five essay prompts were offered tailored to the participants’ CEFR level. Topics included Family and Friends, Sports and Hobbies, Spring Break, Travel Experience, and Weekly Schedule, with each essay suggested to be 120-500 words. 
A user survey was designed for UI and UX assessment, using a 5-point Likert scale with one-choice answers, along with two open-ended questions regarding the most and least useful features. 
Participants had 20 minutes for writing, 10 for corrections, and 10 for a user survey. A1 participants prepared texts in advance, allowing more time for correction.

\hidden{
Data collection is structured around tracking the evolution of each participant's writing. The system records the entire essay submission history, but only those with multiple revisions are included for analysis, as single submissions are excluded from further consideration. The focus is on submissions that demonstrate incremental changes relative to their predecessors; essays with drastic differences from the previous version are omitted. Evaluation metrics include the count of grammar errors (sourced from the GED/C module) and the CEFR level assigned by the AES module.

\begin{table}[ht]
  \centering
  \begin{tabular}{lcc}
    \hline
    \textbf{Category}    & \textbf{Grammar} & \textbf{CEFR Score} \\
    \hline
    Improvements         & \textbf{8}              & 0             \\
    Deterioration        & 1              & 0             \\
    \hline
  \end{tabular}
  \caption{Out of 12 essays with multiple submissions showing gradual changes, 8 cases demonstrated grammar improvements and 1 case showed grammar deterioration, with no changes in score observed.}
  \label{tab:essay_changes}
\end{table}
}

Out of 112 total submissions, where users clicked the Check button and received feedback, 67 submissions were selected, representing the work of 12 different users, because they provided multiple submissions to incremental improvements to a single essay. 8 of these users reduced errors in their essay. One user submission contained only 3 errors in a 212-word initial draft but 4 errors in the final version, but with high CEFR scores suggesting this participant focused on content rather than error correction. The remaining submissions were by A1-B1 learners, where submissions typically contained tens of grammar errors.

No instances of overall CEFR score improvement were observed during the 30-minute writing sessions. Significant score improvements on this relatively course-grained scale would likely require a much longer learning period.


\begin{table}[t!]
  \centering
  \begin{tabular}{lcc}
    \hline
    \textbf{Criteria} & \textbf{Avg. Score} & \textbf{Std. Dev} \\
    \hline
    Clear navigation     & 3.68 & 0.90 \\
    User-friendly        & 3.71 & 0.89 \\
    Intuitive            & 3.59 & 1.09 \\
    Visually Appealing   & 3.03 & 1.03 \\
    Overall Satisfaction & 3.65 & 0.58 \\
    \hline
  \end{tabular}
  \caption{User feedback survey ratings regarding the UI experience. Ratings are on a 5-point Likert scale, with 5 being strongly positive, 3 neutral, and 1 strongly negative.}
  \label{tab:user-feedback}
\end{table}
%

\hidden{
However, due to the imposed time limit and challenges related to confident Arabic typing skills, many users were unable to eliminate all errors during the session. This suggests that while the tool supports effective error correction, further improvements in interface efficiency or additional practice opportunities may be required to help users fully refine their writing within the given time constraints.
}

The survey results shown in Table~\ref{tab:user-feedback} indicate that the overall user experience of the system is moderately positive (see Appendix \ref{appendix2} for more details). Criteria such as ``\textit{Clear navigation}'', ``\textit{User-friendly}'', and ``\textit{Overall Satisfaction}'' all received average scores around 3.65 to 3.71, suggesting that users generally find {\arwi} easy to navigate and use. However, the ``\textit{Visually Appealing}'' criterion received a lower average score of 3.03, indicating room for improvement in visual design.
Standard deviations (0.58 to 1.09) show a moderate degree of variability in users’ perceptions, with the ``Intuitive'' rating exhibiting slightly higher deviation. This suggests that while many users appreciate the UI's intuitiveness, there is a subset for whom it is less clear.
When asked whether they would recommend the system to others, approximately 85\% of users responded affirmatively.

\section{Conclusions and Outlook}
\label{sec:conclusion}
By integrating a collection of essay prompts, a text editor, grammar error detection, correction suggestions, and automated essay scoring modules, {\arwi} provides targeted, iterative, actionable feedback that allows users to improve their writing and see improvements in their writing quality over time. We make {\arwi} publicly available at: \url{https://arwi.mbzuai.ac.ae/}. 

Our preliminary experiment suggests the system is useful, but improvements are needed to the UI, a more fine-grained representation of progress would be useful, and more intuitive error correction hints are needed. We intend to incrementally improve the system based on further user experimentation, feedback, and analytics.

\hidden{
Another contribution of this work is the creation of an infrastructure for building a truly diverse Arabic grammar corpus that supports learners at various proficiency levels, encompasses a wide array of topics along with style nuances. The empirical evidence and user feedback presented herein affirm the tool’s effectiveness, paving the way for further innovations in Arabic NLP and educational technology while addressing longstanding gaps in Arabic writing assistance.
}

\section*{Limitations}

Several aspects of {\arwi} require further refinement. The user interface needs adjustments based on user study feedback, including font size and screen real estate optimization. Error detection, classification, and correction suggestions require improved accuracy. Additionally, a larger study with a more diverse pool of native and non-native students across age groups, along with teacher feedback, is essential for a more comprehensive evaluation.

\section*{Ethical Considerations}
The study parameters were approved by the internal review board (IRB) of our university. All user study participants were volunteers, and the purpose of the study was explained to them directly. 

We recognize that AI assessment systems can make errors that may impact the student learning process and could be misused. This is not our intention.  {\arwi} is designed to serve as a support tool for teachers and learners, not as a standalone evaluator.

\section*{Acknowledgments}
We thank the students in the Arabic Studies program at New York University Abu Dhabi for their enthusiastic participation in our study. We also thank Khulood Kittaneh and Mohammed Muqbel for their valuable support in coordinating the study.





\bibliography{anthology, custom}

\begin{thebibliography}{29}
\providecommand{\natexlab}[1]{#1}

\bibitem[{Alfaifi et~al.(2013)Alfaifi, Atwell, and Abuhakema}]{alfaifi2013error}
A.~Alfaifi, E.~Atwell, and G.~Abuhakema. 2013.
\newblock \href {https://doi.org/10.1007/978-3-642-40722-2_2} {Error annotation of the {A}rabic learner corpus}.
\newblock In \emph{Language Processing and Knowledge in the Web}, volume 8105 of \emph{Lecture Notes in Computer Science}. Springer, Berlin, Heidelberg.

\bibitem[{Alfaifi and Atwell(2012)}]{alfaifi2012arabic}
Abdullah Alfaifi and Eric Atwell. 2012.
\newblock Arabic learner corpora (alc): A taxonomy of coding errors.
\newblock In \emph{The 8th International Computing Conference in Arabic}.

\bibitem[{Alhafni and Habash(2025)}]{alhafni:2025:enhancing}
Bashar Alhafni and Nizar Habash. 2025.
\newblock \href {https://arxiv.org/abs/2503.00985} {Enhancing text editing for grammatical error correction: Arabic as a case study}.
\newblock \emph{Preprint}, arXiv:2503.00985.

\bibitem[{Alhafni et~al.(2023)Alhafni, Inoue, Khairallah, and Habash}]{alhafni-etal-2023-advancements}
Bashar Alhafni, Go~Inoue, Christian Khairallah, and Nizar Habash. 2023.
\newblock \href {https://doi.org/10.18653/v1/2023.emnlp-main.396} {Advancements in {A}rabic grammatical error detection and correction: An empirical investigation}.
\newblock In \emph{Proceedings of the 2023 Conference on Empirical Methods in Natural Language Processing}, pages 6430--6448, Singapore. Association for Computational Linguistics.

\bibitem[{Belkebir and Habash(2021)}]{belkebir-habash-2021-automatic}
Riadh Belkebir and Nizar Habash. 2021.
\newblock \href {https://doi.org/10.18653/v1/2021.conll-1.47} {Automatic error type annotation for {A}rabic}.
\newblock In \emph{Proceedings of the 25th Conference on Computational Natural Language Learning}, pages 596--606, Online. Association for Computational Linguistics.

\bibitem[{Bryant et~al.(2019)Bryant, Felice, Andersen, and Briscoe}]{bryant-etal-2019-bea}
Christopher Bryant, Mariano Felice, {\O}istein~E. Andersen, and Ted Briscoe. 2019.
\newblock \href {https://doi.org/10.18653/v1/W19-4406} {The {BEA}-2019 shared task on grammatical error correction}.
\newblock In \emph{Proceedings of the Fourteenth Workshop on Innovative Use of NLP for Building Educational Applications}, pages 52--75, Florence, Italy. Association for Computational Linguistics.

\bibitem[{Council~of Europe(2001)}]{cefr2001}
C.~o.~E. Council~of Europe. 2001.
\newblock Common european framework of reference for languages: learning, teaching, assessment.

\bibitem[{Dahlmeier et~al.(2013)Dahlmeier, Ng, and Wu}]{dahlmeier-etal-2013-building}
Daniel Dahlmeier, Hwee~Tou Ng, and Siew~Mei Wu. 2013.
\newblock \href {https://aclanthology.org/W13-1703/} {Building a large annotated corpus of learner {E}nglish: The {NUS} corpus of learner {E}nglish}.
\newblock In \emph{Proceedings of the Eighth Workshop on Innovative Use of {NLP} for Building Educational Applications}, pages 22--31, Atlanta, Georgia. Association for Computational Linguistics.

\bibitem[{Fitria(2023)}]{fitria2023artificial}
Tira~Nur Fitria. 2023.
\newblock Artificial intelligence (ai) technology in openai chatgpt application: A review of chatgpt in writing {E}nglish essay.
\newblock In \emph{ELT Forum: Journal of English Language Teaching}, volume~12, pages 44--58.

\bibitem[{Granger(1998)}]{granger1998}
Sylviane Granger. 1998.
\newblock The computer learner corpus: A versatile new source of data for sla research.
\newblock In Sylviane Granger, editor, \emph{Learner English on Computer}, pages 3--18. Addison Wesley Longman, London \& New York.

\bibitem[{Habash and Palfreyman(2022)}]{habash-palfreyman-2022-zaebuc}
Nizar Habash and David Palfreyman. 2022.
\newblock \href {https://aclanthology.org/2022.lrec-1.9/} {{ZAEBUC}: An annotated {A}rabic-{E}nglish bilingual writer corpus}.
\newblock In \emph{Proceedings of the Thirteenth Language Resources and Evaluation Conference}, pages 79--88, Marseille, France. European Language Resources Association.

\bibitem[{Inoue et~al.(2021)Inoue, Alhafni, Baimukan, Bouamor, and Habash}]{inoue-etal-2021-interplay}
Go~Inoue, Bashar Alhafni, Nurpeiis Baimukan, Houda Bouamor, and Nizar Habash. 2021.
\newblock \href {https://aclanthology.org/2021.wanlp-1.10/} {The interplay of variant, size, and task type in {A}rabic pre-trained language models}.
\newblock In \emph{Proceedings of the Sixth Arabic Natural Language Processing Workshop}, pages 92--104, Kyiv, Ukraine (Virtual). Association for Computational Linguistics.

\bibitem[{Kim et~al.(2024)Kim, Kim, Lee, Yoon, Myung, Yoo, Lim, Han, Kim, Ahn, Kim, Oh, Hong, and Lee}]{kim2024designingpromptanalyticsdashboards}
Minsun Kim, SeonGyeom Kim, Suyoun Lee, Yoosang Yoon, Junho Myung, Haneul Yoo, Hyunseung Lim, Jieun Han, Yoonsu Kim, So-Yeon Ahn, Juho Kim, Alice Oh, Hwajung Hong, and Tak~Yeon Lee. 2024.
\newblock \href {https://arxiv.org/abs/2405.19691} {Designing prompt analytics dashboards to analyze student-chatgpt interactions in efl writing}.
\newblock \emph{Preprint}, arXiv:2405.19691.

\bibitem[{Mohit et~al.(2014)Mohit, Rozovskaya, Habash, Zaghouani, and Obeid}]{mohit-etal-2014-first}
Behrang Mohit, Alla Rozovskaya, Nizar Habash, Wajdi Zaghouani, and Ossama Obeid. 2014.
\newblock \href {https://doi.org/10.3115/v1/W14-3605} {The first {QALB} shared task on automatic text correction for {A}rabic}.
\newblock In \emph{Proceedings of the {EMNLP} 2014 Workshop on {A}rabic Natural Language Processing ({ANLP})}, pages 39--47, Doha, Qatar. Association for Computational Linguistics.

\bibitem[{Napoles et~al.(2019)Napoles, N{\u{a}}dejde, and Tetreault}]{napoles-etal-2019-enabling}
Courtney Napoles, Maria N{\u{a}}dejde, and Joel Tetreault. 2019.
\newblock \href {https://doi.org/10.1162/tacl_a_00282} {Enabling robust grammatical error correction in new domains: Data sets, metrics, and analyses}.
\newblock \emph{Transactions of the Association for Computational Linguistics}, 7:551--566.

\bibitem[{Napoles et~al.(2017)Napoles, Sakaguchi, and Tetreault}]{napoles-sakaguchi-tetreault:2017:EACLshort}
Courtney Napoles, Keisuke Sakaguchi, and Joel Tetreault. 2017.
\newblock \href {http://www.aclweb.org/anthology/E17-2037} {Jfleg: A fluency corpus and benchmark for grammatical error correction}.
\newblock In \emph{Proceedings of the 15th Conference of the European Chapter of the Association for Computational Linguistics: Volume 2, Short Papers}, pages 229--234, Valencia, Spain. Association for Computational Linguistics.

\bibitem[{Ng et~al.(2014)Ng, Wu, Briscoe, Hadiwinoto, Susanto, and Bryant}]{ng-etal-2014-conll}
Hwee~Tou Ng, Siew~Mei Wu, Ted Briscoe, Christian Hadiwinoto, Raymond~Hendy Susanto, and Christopher Bryant. 2014.
\newblock \href {https://doi.org/10.3115/v1/W14-1701} {The {C}o{NLL}-2014 shared task on grammatical error correction}.
\newblock In \emph{Proceedings of the Eighteenth Conference on Computational Natural Language Learning: Shared Task}, pages 1--14, Baltimore, Maryland. Association for Computational Linguistics.

\bibitem[{Nicholls et~al.(2024)Nicholls, Caines, and Buttery}]{wicorpus24}
Diane Nicholls, Andrew Caines, and Paula Buttery. 2024.
\newblock \href {https://doi.org/10.17863/CAM.112997} {The {W}rite \& {I}mprove {C}orpus 2024: Error-annotated and {CEFR}-labelled essays by learners of {E}nglish}.
\newblock \emph{Research Outputs}.

\bibitem[{Omelianchuk et~al.(2024)Omelianchuk, Liubonko, Skurzhanskyi, Chernodub, Korniienko, and Samokhin}]{omelianchuk2024pillarsgrammaticalerrorcorrection}
Kostiantyn Omelianchuk, Andrii Liubonko, Oleksandr Skurzhanskyi, Artem Chernodub, Oleksandr Korniienko, and Igor Samokhin. 2024.
\newblock \href {https://arxiv.org/abs/2404.14914} {Pillars of grammatical error correction: Comprehensive inspection of contemporary approaches in the era of large language models}.
\newblock \emph{Preprint}, arXiv:2404.14914.

\bibitem[{Pack et~al.(2024)Pack, Barrett, and Escalante}]{pack2024large}
Austin Pack, Alex Barrett, and Juan Escalante. 2024.
\newblock Large language models and automated essay scoring of {E}nglish language learner writing: Insights into validity and reliability.
\newblock \emph{Computers and Education: Artificial Intelligence}, 6:100234.

\bibitem[{Qwaider et~al.(2025)Qwaider, Alhafni, Chirkunov, Habash, and Briscoe}]{qwaider2025enhancingarabicautomatedessay}
Chatrine Qwaider, Bashar Alhafni, Kirill Chirkunov, Nizar Habash, and Ted Briscoe. 2025.
\newblock \href {https://arxiv.org/abs/2503.17739} {Enhancing arabic automated essay scoring with synthetic data and error injection}.
\newblock \emph{Preprint}, arXiv:2503.17739.

\bibitem[{Rozovskaya et~al.(2015)Rozovskaya, Bouamor, Habash, Zaghouani, Obeid, and Mohit}]{rozovskaya-etal-2015-second}
Alla Rozovskaya, Houda Bouamor, Nizar Habash, Wajdi Zaghouani, Ossama Obeid, and Behrang Mohit. 2015.
\newblock \href {https://doi.org/10.18653/v1/W15-3204} {The second {QALB} shared task on automatic text correction for {A}rabic}.
\newblock In \emph{Proceedings of the Second Workshop on {A}rabic Natural Language Processing}, pages 26--35, Beijing, China. Association for Computational Linguistics.

\bibitem[{Ryding and Wilmsen(2021)}]{Ryding_Wilmsen_2021}
Karin Ryding and David Wilmsen, editors. 2021.
\newblock \emph{The Cambridge Handbook of Arabic Linguistics}.
\newblock Cambridge Handbooks in Language and Linguistics. Cambridge University Press.

\bibitem[{Sanz-Valdivieso(2024)}]{writing2024tools}
Lucía Sanz-Valdivieso. 2024.
\newblock \href {https://doi.org/10.1109/TPC.2024.3419288} {Technology-powered multilingual professional and technical writing: An integrative literature review of landmark and the latest writing assistance tools}.
\newblock \emph{IEEE Transactions on Professional Communication}, 67(3):301--315.

\bibitem[{Sengupta et~al.(2023)Sengupta, Sahu, Jia, and et~al.}]{sengupta2023jais}
Neha Sengupta, Sunil~Kumar Sahu, Bokang Jia, and et~al. 2023.
\newblock \href {https://arxiv.org/abs/2308.16149} {Jais and jais-chat: Arabic-centric foundation and instruction-tuned open generative large language models}.
\newblock \emph{Preprint}, arXiv:2308.16149.

\bibitem[{Team et~al.(2025)Team, Abbas, Ahmad, and et~al.}]{team2025fanar}
Fanar Team, Ummar Abbas, Mohammad~Shahmeer Ahmad, and et~al. 2025.
\newblock Fanar: An arabic-centric multimodal generative ai platform.
\newblock \emph{arXiv preprint arXiv:2501.13944}.

\bibitem[{{United Nations}(2024)}]{UNPage2024}
{United Nations}. 2024.
\newblock \href {https://www.un.org/en/observances/arabiclanguageday} {Arabic language and {AI}: Advancing innovation while preserving cultural heritage}.
\newblock Accessed: 2025-02-28.

\bibitem[{Wu et~al.(2023)Wu, Wang, Wan, Jiao, and Lyu}]{wu2023chatgptgrammarlyevaluatingchatgpt}
Haoran Wu, Wenxuan Wang, Yuxuan Wan, Wenxiang Jiao, and Michael Lyu. 2023.
\newblock \href {https://arxiv.org/abs/2303.13648} {Chatgpt or grammarly? evaluating chatgpt on grammatical error correction benchmark}.
\newblock \emph{Preprint}, arXiv:2303.13648.

\bibitem[{Yancey et~al.(2023)Yancey, LaFlair, Verardi, and Burstein}]{Yancey2023RatingSL}
K.~Yancey, Geoffrey~T. LaFlair, Anthony Verardi, and Jill Burstein. 2023.
\newblock \href {https://api.semanticscholar.org/CorpusID:259376840} {Rating short l2 essays on the cefr scale with gpt-4}.
\newblock In \emph{Workshop on Innovative Use of NLP for Building Educational Applications}.

\end{thebibliography}

\appendix
\newpage
\onecolumn

\section{{\arwi} Interface}
\label{appendix}
\begin{figure*}[ht]
    \centering
    \includegraphics[width=1\linewidth]{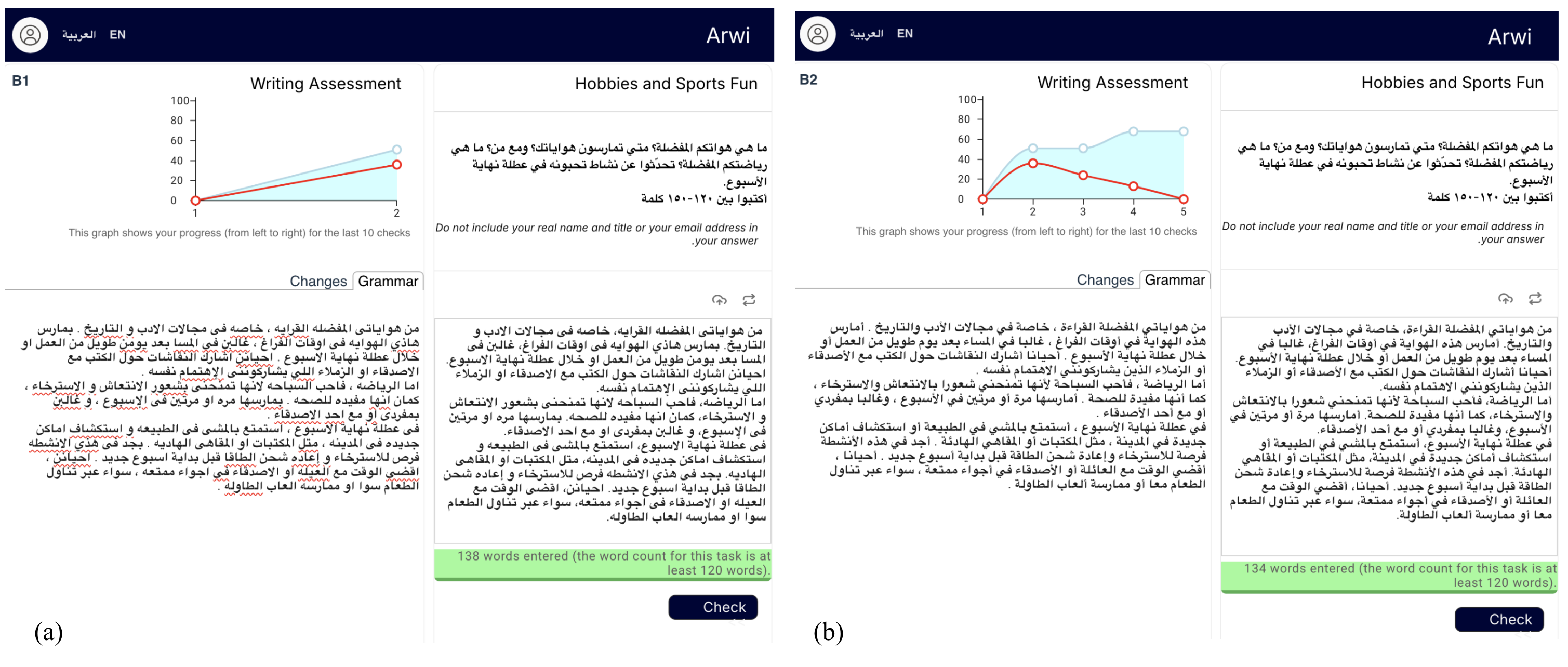}
    \caption{A before-and-after example of using  {\arwi}'s English interface. In (a) the text receives a B1 CEFR and a large number of errors marked with red underlining; in (b) the results shows improved writing and is raised to B2 CEFR.
    The \textbf{essay prompt} can be translated as ``\textit{What are your favorite hobbies? When do you practice your hobbies? And with whom? What is your favorite sport? Talk about an activity you enjoy on the weekend. Write between 120-150 words.}'' The \textbf{written essay} can be translated as: ``\textit{One of my favorite hobbies is reading, especially in the fields of literature and history. I engage in this hobby during my free time, often in the evening after a long day of work or during the weekend. Sometimes, I participate in book discussions with friends or colleagues who share the same interest. As for sports, I enjoy swimming because it gives me a sense of refreshment and relaxation, and it is also beneficial for my health. I practice it once or twice a week, often alone or with a friend. During the weekend, I enjoy walking in nature or exploring new places in the city, such as libraries or quiet cafés. I find these activities to be an opportunity to relax and recharge before the start of a new week. Sometimes, I spend time with family or friends in a fun atmosphere, whether by sharing a meal together or playing board games.}''
    }
    \label{fig:arwi-interface-en}
\end{figure*}

\newpage
\onecolumn
\section{User Feedback Survey}
\label{appendix2}
\begin{figure*}[ht]
    \centering
    \includegraphics[width=1\linewidth]{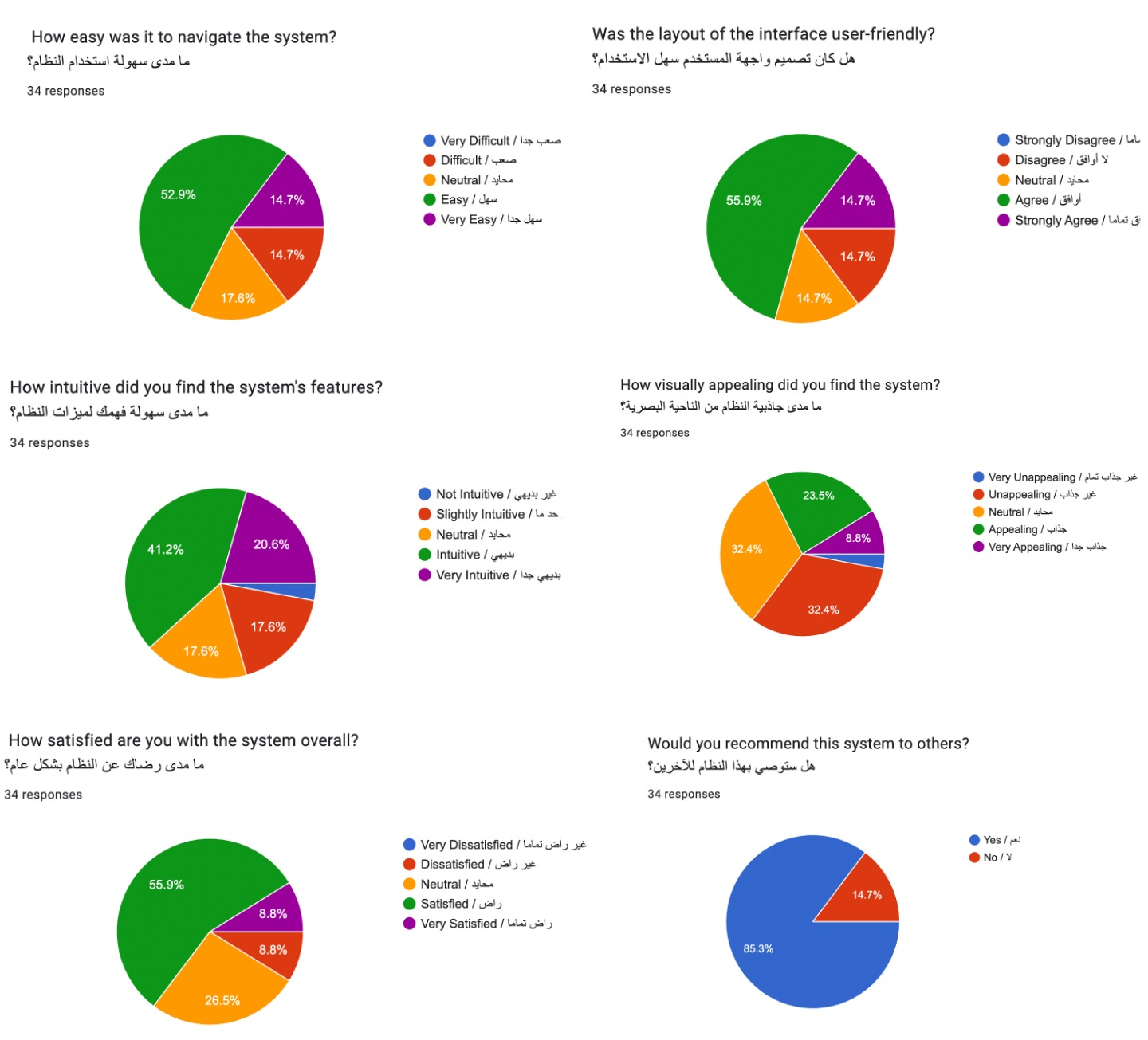}
    \caption{Qualitative feedback collected from 34 users who participated in the preliminary experiments with Arwi. The survey comprised five one-choice questions rated on a 5-point Likert scale and one binary question. The results highlight that certain aspects of the user interface--specifically its intuitiveness and visual appeal--require further refinement. Overall, users provided moderately positive feedback regarding their experience of usage. }
    \label{fig:arwi-user-survey}
\end{figure*}

\end{document}